 \documentclass[pmlr, twocolumn]{jmlr} % W&CP article

\usepackage{booktabs}
\usepackage[load-configurations=version-1]{siunitx} % newer version

\usepackage{caption}
\usepackage{placeins}

\usepackage{tikz-qtree}

\usepackage{tikz}
\usetikzlibrary{shapes,arrows, positioning}
\tikzstyle{block} = [rectangle, draw, fill=gray!20, 
    text width=5em, text centered, rounded corners=0.25em, minimum height=2em]
\tikzstyle{line} = [draw, -latex']

\newcommand{\specialcell}[2][c]{%
  \begin{tabular}[#1]{@{}r@{}}#2\end{tabular}}
\newcommand{\specialcellleft}[2][c]{%
  \begin{tabular}[#1]{@{}l@{}}#2\end{tabular}}

\jmlrvolume{ML4H Extended Abstract Arxiv Index}
\jmlryear{2020}
\jmlrsubmitted{2020}
\jmlrpublished{}
\jmlrpages{}
\jmlrworkshop{Machine Learning for Health (ML4H) at NeurIPS 2020} 

\title[Towards Automated Anamnesis Summarization]{Towards Automated Anamnesis Summarization: BERT-based Models for Symptom Extraction}

\usepackage{amsmath}

\author{%
    \Name{Anton Schäfer}\footnotemark[1] \Email{scanton@ethz.ch}\\
    \Name{Nils Blach}\footnotemark[1] \Email{blachn@ethz.ch}\\
    \Name{Oliver Rausch}\footnotemark[1] \phantom{\thanks{Equal contribution}} \Email{rauscho@ethz.ch}\\
    \addr ETH Zurich
    \AND
    \Name{Maximilian Warm} \Email{maximilian.warm@med.uni-muenchen.de}\\
    \Name{Nils Krüger} \Email{nils.krueger@med.uni-muenchen.de}\\
    \addr LMU Munich
}

\begin{document}

\maketitle

\begin{abstract}
Professionals in modern healthcare systems are increasingly burdened by documentation workloads. Documentation of the initial patient anamnesis is particularly relevant, forming the basis of successful further diagnostic measures. However, manually prepared notes are inherently unstructured and often incomplete.
In this paper, we investigate the potential of modern NLP techniques to support doctors in this matter. We present a dataset of German patient monologues, and formulate a well-defined information extraction task under the constraints of real-world utility and practicality.
In addition, we propose BERT-based models in order to solve said task. We can demonstrate promising performance of the models in both symptom identification and symptom attribute extraction, significantly outperforming simpler baselines.% , and attaining F1 scores of up to $90\%$ and $70\%$ respectively.
\end{abstract}

\section{Introduction \& Related Work}

Thorough documentation of a patient's clinical encounters remains a highly sought goal in modern healthcare systems.
Especially of importance are the patient's initial complaints, which often guide diagnostic measures.
These are usually delineated in a patient monologue during an initial anamnesis interview between physician and patient.
During such conversations, documentation is often recorded in handwritten or typed form. This is very time effective, but usually results in brief, non-standardized, and potentially incomplete notes which are difficult to query or investigate. The professional is often forced to trade off documentation completeness, patient attention and time expenditure.
A system that automatically extracts relevant information from such monologues could greatly improve documentation completeness and time efficiency.

Recent methods achieve promising results for neural abstractive summarization (\citealt{shi2018neural}). However, such sequence to sequence approaches usually require large amounts of training data, which is particularly hard to obtain in the medical domain (\citealt{quiroz2019challenges}).

Other work investigates symptom extraction from clinical text. \cite{jackson2017natural} extract symptoms from discharge notes, and others classify chief complaints from Electronic Health Records (\citealt{osti_1637127}, \citealt{Chapman2005ClassifyingFT}).
In the conversational domain, \cite{liu-etal-2019-fast} identify attributes of symptoms on a corpus of English post-discharge nurse-patient dialogues.

Recently, the transformer-based (\citealt{vaswani2017attention}) BERT model (\citealt{devlin-etal-2019-bert}) attained state of the art results on various NLP tasks. The approach has also been applied in medical informatics, yielding domain-specific models such as Clinical BERT (\citealt{alsentzer-etal-2019-publicly}) that have been successfully applied to medical information extraction (\citealt{10.1093/jamia/ocz096}, \citealt{10.1093/jamiaopen/ooaa022}).
In this paper, we focus on a dataset of German patient monologues. Unfortunately, there are much fewer resources available for the German language; e.g. no domain specific pretrained BERT models exist, and  publicly available medical data is scarce. This significantly increases the difficulty of creating effective models when dealing with German text.

Our goal in this work is to extract all symptoms out of German patient monologues. Our contributions are threefold:
i) we present a dataset that consists of written descriptions of patient conditions, ii) we propose a pragmatically grounded task in which each description is annotated with a structured summary of reported symptoms and their attributes, and iii) we present BERT-based models to solve the task.

\section{Dataset}
We collect publicly accessible text-based patient written condition descriptions (henceforth denoted as ``post'') from German medical forums. We select posts that describe gastrointestinal conditions.

After conducting a survey of active medical professionals in Germany, we find that extracting concise symptom information from the often non-technical and verbose patient descriptions is most relevant in the anamnesis. Although the history of treatment, condition, and medication are also useful, we focus on symptom extraction as a first step.

For each forum post, we label the described symptoms according to a hierarchical collection of common symptoms. Similar to \cite{liu-etal-2019-fast}, we further annotate attributes that convey additional information, categorized as one of several categories: location, description, time, frequency and action. The survey participants confirm that this serves as a useful summary.
We formulate rigorous rules to reduce room for interpretation. To ensure consistency, we label a large portion of the posts twice and merge the results in a final review.

In total, the dataset consists of $125$ unique posts, totaling $592$ symptoms and $1276$ attributes, of which $729$ were unique post segments. Further details and dataset statistics can be found in Appendix \ref{datasetdetails}.
\section{Methods}
To infer the labelings, we divide the task into two stages: \emph{Symptom Classification} and \emph{Attribute Extraction}. In both stages, we make use of a pretrained German BERT model (\citealt{devlin-etal-2019-bert}) from HuggingFace's Transformers (\citealt{Wolf2019HuggingFacesTS}): we use \texttt{bert-base-german-cased} provided by \texttt{deepset.ai} for all experiments.

\subsection{Symptom Classification}
\paragraph{Baseline Models.}
We present two baseline models. The first is an MLP with a sigmoid output layer, trained on TF-IDF (\citealt{Ramos2003UsingTT}) features. The second is a pretrained BERT model followed by a sigmoid classification layer. For both models, the dimension of the output layer corresponds to the number of output symptoms in the training set.
\paragraph{BERT Symptom Query Model.}
The baseline classifiers can naturally only predict symptoms that occur in the training dataset. We propose a symptom query model which is able to generalize to unseen symptoms. We feed the symptom description -- a short, layman description of the symptom -- along with the patient written text into the model, and attempt to predict whether the symptom is present in the text or not.
Due to the large amount of possible negative samples, we randomly select one negative sample for each positive sample in a post.

\paragraph{Curriculum Learning.}
To gain further performance, we employ a curriculum learning scheme (\citealt{bengio2009curriculum}). The hierarchical nature of the symptom collection allows us to form negative samples which are gradually more complex to classify. For instance, symptoms which are close in the hierarchy are more difficult to differentiate.

\paragraph{Augmented Descriptions.}
To improve generalization ability to new symptoms, we extract all text segments in the corpus that were annotated as symptoms and introduce them as alternative layman descriptions of the respective symptom during training.
This approach yields an order of magnitude increase in the amount of positive samples.

\subsection{Attribute Extraction.}
To extract the relevant attributes for a given symptom we adopt a question answering (QA) style approach. The model receives the post and the symptom description as input and predicts which text sections constitute attributes.
Unlike in many QA tasks, (e.g. \citealt{rajpurkar-etal-2016-squad}, \citealt{liu-etal-2019-fast}), we often require multiple answers to a single question, as a symptom can have multiple attributes of the same type. 
We introduce two methods to solve this task. For each of the methods, we train six models; a separate model for each of the five attribute types, and a ``general'' model that is trained to predict all attributes at once (see Appendix \ref{app:models:extraction}).

\paragraph{Start-End Prediction.}
\label{start_end}
Conceptually similar to the QA model described by \cite{devlin-etal-2019-bert}, we combine BERT with two final shared linear layers applied to all tokens. For each token, we predict if it is the start or the end of an attribute.

To account for the multiple possible answers, we use the sigmoid activation function and consider multiple start-end pairs.  We often observe very long predictions due to mismatched start and end tokens. This occurs when two attributes are predicted correctly, but the start of one attribute is matched with the end of the other, resulting in a very long token sequence. To combat this, we employ a heuristic post-processing method to form the final predictions (see Appendix \ref{postprocessing}).

\paragraph{Contiguous Prediction.}
To circumvent mismatching start and end tokens, we propose to consider each token separately. We append a linear layer after the BERT model, with which we independently classify whether each token is part of a given attribute. We then construct the predictions as all longest sequences of tokens that all have predicted probabilities greater than a threshold of $0.7$.
\begin{table}[b!]
    \begin{tabular}{lrrr}
        \toprule
        Method             & F1               & Prec.   & Rec.    \\
        \midrule
        TF-IDF + MLP       & $0.652$          & $0.768$ & $0.566$ \\
        BERT + Sigmoid & $0.672$          & $0.857$ & $0.553$ \\
        \hline
        $\text{BERT}_\text{SQ}$            & $0.838$          & $0.861$ & $0.816$ \\
        $\text{BERT}_\text{SQ}$ CL         & $0.844$          & $0.873$ & $0.816$ \\
        $\text{BERT}_\text{SQ}$ AD         & $0.857$          & $0.846$ & $0.868$ \\
        $\text{BERT}_\text{SQ}$ CL+AD      & $\mathbf{0.903}$ & $0.956$ & $0.855$ \\
        \bottomrule  
    \end{tabular}
    \centering
    \caption{The F1 score, precision and recall of different methods on the Symptom Classification task. We denote the BERT Symptom Query model as $\text{BERT}_\text{SQ}$, the curriculum learning scheme as CL, and the augmented dataset as AD.}
    \label{symptom_classification_results}

\end{table}

\begin{table*}[t!]
\begin{tabular}{lrrrrr}
        \toprule
        Method            & Location            & Description       & Time           & Frequency      & Action        \\
        \midrule
        $\text{BERT}_\text{start\_end}$         & $0.56$            & $0.25$            & $\mathbf{0.42}$   &  $0.19 $          & $0.12$\\
        $\text{BERT}_\text{start\_end}$ general & $\mathbf{0.70}$   & $0.42$            & $0.29$            & $0.19$            & $0.27$\\
        $\text{BERT}_\text{contiguous}$         & $0.63$            & $\mathbf{0.43}$   & $0.31$            & $0.35$            & $0.21$\\
        $\text{BERT}_\text{contiguous}$ general & $0.66$            & $0.35$            & $0.34$            & $\mathbf{0.42}$   & $\mathbf{0.32}$  \\
        \bottomrule
    \end{tabular}
    \centering
    \caption{Attribute extraction results on the held out test set. We report the token-wise F1 score based on the constructed predictions. For precision and recall see Appendix \ref{app:more_results}.}
    \label{attribute_extraction_results}
\end{table*}
\FloatBarrier
\section{Experiments}

The dataset is split into train, test and validation sets. The test set is formed using $20\%$ of all labelings which were identified as correct after double labeling; we observe that these labelings are of higher quality. The validation set is formed using $10\%$ of all remaining labelings. We evaluate the model that achieved the highest validation F1 score (sum 
of F1 scores over all attributes for the general attribute extraction models) and report it's scores on the held out test set. All F1 scores are micro averaged across all classes. All models are implemented using PyTorch (\citealt{paszke2019pytorch}).

\section{Results}
\paragraph{Symptom Classification.}
We observe that the symptom query model architecture outperforms both the TF-IDF + MLP and the sigmoid classifier baselines. Both curriculum learning and data augmentation appear to be beneficial and together yield a performance increase of $0.065$ F1 compared to the base BERT Symptom Query model (see Table \ref{symptom_classification_results}).

We further notice that the symptom query model with augmented descriptions generalizes better to unseen symptoms, as illustrated in Appendix \ref{app:generalization}.

\paragraph{Attribute Extraction.}
We report results by attribute type in Table \ref{attribute_extraction_results}. Both methods appear to produce useful predictions (see Appendix \ref{app:example_preds}). We observe that the ``general'' models that are trained on all attributes at once outperform the respective separately trained counterparts on most attributes.

The location attribute is predicted particularly well by almost all models, with F1 scores up to 0.7. We hypothesize that attributing a location to the correct symptom is simpler than correctly matching the other, more generic attributes: ``head'' is likely to be an attribute to ``headache'', while e.g. ``rarely'' could be a frequency attribute to almost any symptom.

\section{Conclusion}
In this work, we propose an approach to extract symptom information under the constraints of real-world utility and practicality. To the best of our knowledge, we are the first to attempt such a task in German. We demonstrate promising performance even on a relatively small dataset using a data augmentation technique and a curriculum learning scheme.

Although we observe limitations in model performance, particularly in the attribute extraction task, we are optimistic that this can be improved with a larger dataset. Further work could investigate the extraction of other relevant information such as pre-existing conditions and medication. With such improvements, we see potential for practical, automated anamnesis summarization in the near future.
\pagebreak
 \acks{
 We would like to thank Konstantinos Dimitriadis for the valuable domain insights and fruitful discussions.
 We are further grateful for Elliot Ash's support and the provided access to the Leonhard Cluster, as well as Julia Vogt's valuable feedback that improved the quality of the paper.
 }

\bibliography{bibliography}

\begin{thebibliography}{17}
\providecommand{\natexlab}[1]{#1}
\providecommand{\url}[1]{\texttt{#1}}
\expandafter\ifx\csname urlstyle\endcsname\relax
  \providecommand{\doi}[1]{doi: #1}\else
  \providecommand{\doi}{doi: \begingroup \urlstyle{rm}\Url}\fi

\bibitem[Alsentzer et~al.(2019)Alsentzer, Murphy, Boag, Weng, Jindi, Naumann,
  and McDermott]{alsentzer-etal-2019-publicly}
Emily Alsentzer, John Murphy, William Boag, Wei-Hung Weng, Di~Jindi, Tristan
  Naumann, and Matthew McDermott.
\newblock Publicly available clinical {BERT} embeddings.
\newblock In \emph{Clinical Natural Language Processing Workshop}. Association
  for Computational Linguistics, June 2019.

\bibitem[Bengio et~al.(2009)Bengio, Louradour, Collobert, and
  Weston]{bengio2009curriculum}
Yoshua Bengio, J{\'e}r{\^o}me Louradour, Ronan Collobert, and Jason Weston.
\newblock Curriculum learning.
\newblock In \emph{International Conference on Machine Learning}, June 2009.

\bibitem[Chang et~al.(2020)Chang, Hong, and Taylor]{10.1093/jamiaopen/ooaa022}
David Chang, Woo~Suk Hong, and Richard~Andrew Taylor.
\newblock {Generating contextual embeddings for emergency department chief
  complaints}.
\newblock \emph{Journal of the American Medical Informatics Association}, July
  2020.

\bibitem[Chapman et~al.(2005)Chapman, Christensen, Wagner, Haug, Ivanov,
  Dowling, and Olszewski]{Chapman2005ClassifyingFT}
W.~Chapman, L.~M. Christensen, M.~Wagner, P.~Haug, O.~Ivanov, J.~Dowling, and
  R.~Olszewski.
\newblock Classifying free-text triage chief complaints into syndromic
  categories with natural language processing.
\newblock \emph{Artificial intelligence in medicine}, January 2005.

\bibitem[Devlin et~al.(2019)Devlin, Chang, Lee, and
  Toutanova]{devlin-etal-2019-bert}
Jacob Devlin, Ming-Wei Chang, Kenton Lee, and Kristina Toutanova.
\newblock {BERT}: Pre-training of deep bidirectional transformers for language
  understanding.
\newblock In \emph{Conference of the North {A}merican Chapter of the
  Association for Computational Linguistics: Human Language Technologies}.
  Association for Computational Linguistics, June 2019.

\bibitem[Jackson et~al.(2017)Jackson, Patel, Jayatilleke, Kolliakou, Ball,
  Gorrell, Roberts, Dobson, and Stewart]{jackson2017natural}
Richard~G Jackson, Rashmi Patel, Nishamali Jayatilleke, Anna Kolliakou, Michael
  Ball, Genevieve Gorrell, Angus Roberts, Richard~J Dobson, and Robert Stewart.
\newblock Natural language processing to extract symptoms of severe mental
  illness from clinical text: the clinical record interactive search
  comprehensive data extraction (cris-code) project.
\newblock \emph{BMJ open}, 2017.

\bibitem[Kingma and Ba(2015)]{DBLP:journals/corr/KingmaB14}
Diederik~P. Kingma and Jimmy Ba.
\newblock Adam: {A} method for stochastic optimization.
\newblock In \emph{International Conference on Learning Representations}, 2015.

\bibitem[Lee et~al.(2019)Lee, Levin, Finley, and Heilig]{osti_1637127}
Scott~H. Lee, Drew Levin, Patrick~D. Finley, and Charles~M. Heilig.
\newblock Chief complaint classification with recurrent neural networks.
\newblock \emph{Journal of Biomedical Informatics}, May 2019.

\bibitem[Liu et~al.(2019)Liu, Lim, Suhaimi, Tong, Ong, Ng, Lee, Macdonald,
  Ramasamy, Krishnaswamy, Chow, and Chen]{liu-etal-2019-fast}
Zhengyuan Liu, Hazel Lim, Nur Farah~Ain Suhaimi, Shao~Chuen Tong, Sharon Ong,
  Angela Ng, Sheldon Lee, Michael~R. Macdonald, Savitha Ramasamy, Pavitra
  Krishnaswamy, Wai~Leng Chow, and Nancy~F. Chen.
\newblock Fast prototyping a dialogue comprehension system for nurse-patient
  conversations on symptom monitoring.
\newblock In \emph{Conference of the North {A}merican Chapter of the
  Association for Computational Linguistics: Human Language Technologies}, June
  2019.

\bibitem[Paszke et~al.(2019)Paszke, Gross, Massa, Lerer, Bradbury, Chanan,
  Killeen, Lin, Gimelshein, Antiga, et~al.]{paszke2019pytorch}
Adam Paszke, Sam Gross, Francisco Massa, Adam Lerer, James Bradbury, Gregory
  Chanan, Trevor Killeen, Zeming Lin, Natalia Gimelshein, Luca Antiga, et~al.
\newblock Pytorch: An imperative style, high-performance deep learning library.
\newblock In \emph{Advances in neural information processing systems}, 2019.

\bibitem[Quiroz et~al.(2019)Quiroz, Laranjo, Kocaballi, Berkovsky, Rezazadegan,
  and Coiera]{quiroz2019challenges}
Juan~C Quiroz, Liliana Laranjo, Ahmet~Baki Kocaballi, Shlomo Berkovsky, Dana
  Rezazadegan, and Enrico Coiera.
\newblock Challenges of developing a digital scribe to reduce clinical
  documentation burden.
\newblock \emph{npj Digital Medicine}, 2019.

\bibitem[Rajpurkar et~al.(2016)Rajpurkar, Zhang, Lopyrev, and
  Liang]{rajpurkar-etal-2016-squad}
Pranav Rajpurkar, Jian Zhang, Konstantin Lopyrev, and Percy Liang.
\newblock {SQ}u{AD}: 100,000+ questions for machine comprehension of text.
\newblock In \emph{Conference on Empirical Methods in Natural Language
  Processing}, November 2016.

\bibitem[Ramos(2003)]{Ramos2003UsingTT}
J.~Ramos.
\newblock Using tf-idf to determine word relevance in document queries.
\newblock 2003.

\bibitem[Shi et~al.(2018)Shi, Keneshloo, Ramakrishnan, and
  Reddy]{shi2018neural}
Tian Shi, Yaser Keneshloo, Naren Ramakrishnan, and Chandan~K Reddy.
\newblock Neural abstractive text summarization with sequence-to-sequence
  models.
\newblock \emph{arXiv preprint arXiv:1812.02303}, 2018.

\bibitem[Si et~al.(2019)Si, Wang, Xu, and Roberts]{10.1093/jamia/ocz096}
Yuqi Si, Jingqi Wang, Hua Xu, and Kirk Roberts.
\newblock {Enhancing clinical concept extraction with contextual embeddings}.
\newblock \emph{Journal of the American Medical Informatics Association}, July
  2019.

\bibitem[Vaswani et~al.(2017)Vaswani, Shazeer, Parmar, Uszkoreit, Jones, Gomez,
  Kaiser, and Polosukhin]{vaswani2017attention}
Ashish Vaswani, Noam Shazeer, Niki Parmar, Jakob Uszkoreit, Llion Jones,
  Aidan~N Gomez, {\L}ukasz Kaiser, and Illia Polosukhin.
\newblock Attention is all you need.
\newblock In \emph{Advances in neural information processing systems}, 2017.

\bibitem[Wolf et~al.(2019)Wolf, Debut, Sanh, Chaumond, Delangue, Moi, Cistac,
  Rault, Louf, Funtowicz, and Brew]{Wolf2019HuggingFacesTS}
Thomas Wolf, Lysandre Debut, Victor Sanh, Julien Chaumond, Clement Delangue,
  Anthony Moi, Pierric Cistac, Tim Rault, R'emi Louf, Morgan Funtowicz, and
  Jamie Brew.
\newblock Huggingface's transformers: State-of-the-art natural language
  processing.
\newblock \emph{ArXiv}, abs/1910.03771, 2019.

\end{thebibliography}

\clearpage

\appendix
\onecolumn
\section{Dataset Details}\label{datasetdetails}

\begin{center}
    \begin{tabular}{lrr}
    \toprule
                         & {Total} \\
                         \midrule
    {Diarrhea}            & 44               \\
    {Nausea}               & 40               \\
    {Upp. Abdominal Pain} & 29               \\
    {Flatulence}           & 28               \\
    {Hematochecia}         & 20              \\
    \bottomrule
    \end{tabular}
    \captionof{table}{Symptom statistics of labeled symptoms, showing the most frequent ``leaf'' symptoms in the dataset according to the hierarchical collection of common symptoms.}
\end{center}

\begin{center}
\centering
\begin{tabular}{lrrrr}
\toprule
                     & \specialcell{Total\\Occurrences} & \specialcell{Unique\\Occurrences} & \specialcell{Mean Attribute\\Length} & \specialcell{Attribute Length\\Std Dev} \\\midrule
{Time}        & 406                        & 222                         & 2.75                          & 1.62                         \\
{Description} & 352                        & 233                         & 1.89                          & 1.04                         \\
{Location}    & 216                        & 89                          & 2.66                          & 1.78                         \\
{Frequency}   & 159                        & 79                          & 1.94                         & 1.33                         \\
{Action}      & 143                        & 106                         & 5.13                          & 2.75                         \\
\bottomrule
\end{tabular}
\captionof{table}{Statistics of labeled attributes classes.}
\end{center}

\clearpage
\section{Model Descriptions}\label{appmodels}

\subsection{Symptom Classification} \label{app:models:classification}
Our architecture (see Figure \ref{model_classification}) consists of the pretrained BERT model, followed by a linear layer which takes the concatentation of the mean of all token hidden states, and the pooler output.\footnote{The pooler output is the hidden state of the CLS token, passed through a linear layer} 

\begin{center}
  \begin{tikzpicture}[node distance = 2cm, auto]
    % Place nodes
    \node [block, text width=15em, minimum height= 4 em, fill=yellow!50] (bert) {BERT};
    
    \node [below=1em of bert.south, anchor=north] (inputs) {[CLS] Symptom [SEP] Post Text [SEP]};
    \node [block, above=1em of bert.east|-bert.north,anchor=south east, text width=11.5em, fill=blue!15] (mean) {Mean};
    \node [block, above=1em of bert.west|-bert.north,anchor=south west, text width=2.5em] (linear) {{\scriptsize Linear}};
    \node [block, above=1em of linear, text width=2.5em, anchor=south, fill=green!20] (tanh) {$\tanh$};
    
    \node [block, above=1em of bert.north|-tanh.north, text width=5em, anchor=south] (biglinear) {Linear};
     \node [block, above=1em of biglinear.north, text width=5em, anchor=south, fill=green!20] (sig) {$\sigma$};
    
    \node [above=1em of sig.north, anchor=south] (output) {Prediction};

    \draw[->, draw=none] (inputs) -- (bert) node [midway, align=center,auto=false] {$\dots$};
    
    \draw[->, draw=none] (mean.south |-bert.north) -- (mean.south) node [midway, align=center,auto=false] {$\dots$};

    \draw[->] (linear.south|-bert.north) - ++(0,0.5em) -| ([xshift=-5em]mean.south);

    \draw[->] ([xshift=5em]mean.south |- bert.north) -- ([xshift=5em]mean.south);
    \draw[->] ([xshift=4em]mean.south |- bert.north) -- ([xshift=4em]mean.south);
    \draw[->] ([xshift=-4em]mean.south |- bert.north) -- ([xshift=-4em]mean.south);
    \draw[->] ([xshift=3em]mean.south |- bert.north) -- ([xshift=3em]mean.south);
    \draw[->] ([xshift=-3em]mean.south |- bert.north) -- ([xshift=-3em]mean.south);
    
    \draw[->] ([xshift=5em] inputs.north) -- ([xshift=5em]bert.south);
    \draw[->] ([xshift=4em] inputs.north) -- ([xshift=4em]bert.south);
    \draw[->] ([xshift=-5em] inputs.north) -- ([xshift=-5em]bert.south);
    \draw[->] ([xshift=-4em] inputs.north) -- ([xshift=-4em]bert.south);
    \draw[->] ([xshift=-3em] inputs.north) -- ([xshift=-3em]bert.south);
    \draw[->] ([xshift=3em] inputs.north) -- ([xshift=3em]bert.south);
    \draw[->] ([xshift=6em] inputs.north) -- ([xshift=6em]bert.south);
    \draw[->] ([xshift=-6em] inputs.north) -- ([xshift=-6em]bert.south);
    \draw[->] ([xshift=7em] inputs.north) -- ([xshift=7em]bert.south);
    \draw[->] ([xshift=-7em] inputs.north) -- ([xshift=-7em]bert.south);
    
    \draw[<-] (linear) -- ([yshift=0.5em]linear.south|-bert.north);
    \draw[->] (linear) -- (tanh);
    \draw[<-] [rounded corners=0.25em] ([xshift=-1em]biglinear.south) |- (tanh);
    \draw[->] ([xshift=1em]biglinear.south|-mean.north) -- ([xshift=1em]biglinear.south);
    \draw[->] (biglinear) -- (sig);
    \draw[->] (sig) -- (output);

\end{tikzpicture}

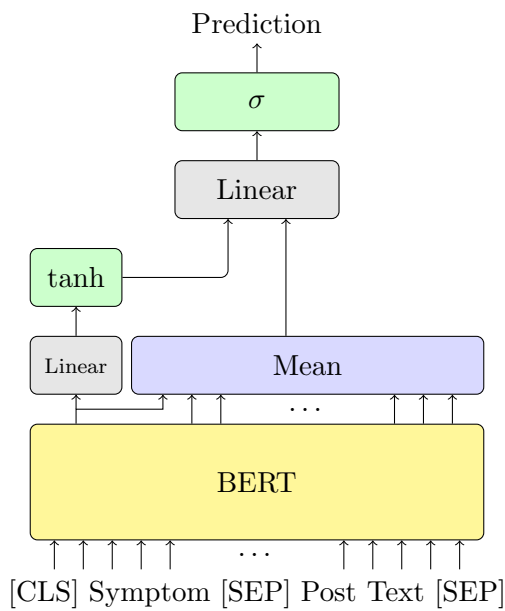
\captionof{figure}{The $\text{BERT}_\text{SQ}$ symptom classification model architecture.}
\label{model_classification}
\end{center}
\subsection{Attribute Extraction} \label{app:models:extraction}
We use the pretrained BERT model, followed by a linear layer that is applied to each token's BERT embedding (see Figure \ref{model_attribute}). The linear layer predicts two probabilities, $p^s$ and $p^e$ for each token (and for each attribute in case of the general models) when using the Start-End method, and a single probability $p^\text{inside}$ per token in case of the Contiguous method (see Table \ref{targets}).

\begin{table}[h!]
\centering
\begin{tabular}{lccccccccc}
\toprule
                  & I & have & pain & in & my & \textbf{right} & \textbf{knee} & and & \textbf{shin} \\
                  \midrule
$p^s$             & 0 & 0    & 0    & 0  & 0  & 1              & 0             & 0   & 1             \\
$p^e$             & 0 & 0    & 0    & 0  & 0  & 0              & 1             & 0   & 1             \\
\midrule
$p^\text{inside}$ & 0 & 0    & 0    & 0  & 0  & 1              & 1             & 0   & 1   \\     
\bottomrule
\end{tabular}
\caption{Example target probabilities for the attribute extraction models. Note that in the actual model, we use wordpiece tokenization, and not word tokenization like in this example.}
\label{targets}
\end{table}
For both methods, we minimize the weighted Negative Log Likelihood loss to ensure equally learning both classes despite the heavily unbalanced dataset.
\\

\begin{center}
  \begin{tikzpicture}[node distance = 2cm, auto]
    % Place nodes
    \node [block, text width=15em, minimum height= 4 em, fill=yellow!50] (bert) {BERT};
    
    \node [below=1em of bert.south, anchor=north] (inputs) {[CLS] Symptom [SEP] Post Text [SEP]};
    \node [block, above=1em of bert.west|-bert.north,anchor=south west, text width=15em, fill=blue!15] (linear) {Linear};
    
     \node [block, above=1em of linear.north, text width=15em, anchor=south, fill=green!20] (sig) {$\sigma$};
    
    \node [above=1em of sig.north, anchor=south] (output) {Predictions};

    \draw[->, draw=none] (inputs) -- (bert) node [midway, align=center,auto=false] {$\dots$};
    
    \draw[->, draw=none] (bert) -- (linear) node [midway, align=center,auto=false] {$\dots$};
    \draw[->, draw=none] (linear) -- (sig) node [midway, align=center,auto=false] {$\dots$};
    \draw[->, draw=none] (sig) -- (output) node [midway, align=center,auto=false] {$\dots$};
    
    \draw[->] ([xshift=5em] inputs.north) -- ([xshift=5em]bert.south);
    \draw[->] ([xshift=4em] inputs.north) -- ([xshift=4em]bert.south);
    \draw[->] ([xshift=-5em] inputs.north) -- ([xshift=-5em]bert.south);
    \draw[->] ([xshift=-4em] inputs.north) -- ([xshift=-4em]bert.south);
    \draw[->] ([xshift=-3em] inputs.north) -- ([xshift=-3em]bert.south);
    \draw[->] ([xshift=3em] inputs.north) -- ([xshift=3em]bert.south);
    \draw[->] ([xshift=6em] inputs.north) -- ([xshift=6em]bert.south);
    \draw[->] ([xshift=-6em] inputs.north) -- ([xshift=-6em]bert.south);
    \draw[->] ([xshift=7em] inputs.north) -- ([xshift=7em]bert.south);
    \draw[->] ([xshift=-7em] inputs.north) -- ([xshift=-7em]bert.south);

    \draw[->] ([xshift=5em] bert.north) -- ([xshift=5em]linear.south);
    \draw[->] ([xshift=4em] bert.north) -- ([xshift=4em]linear.south);
    \draw[->] ([xshift=-5em] bert.north) -- ([xshift=-5em]linear.south);
    \draw[->] ([xshift=-4em] bert.north) -- ([xshift=-4em]linear.south);
    \draw[->] ([xshift=-3em] bert.north) -- ([xshift=-3em]linear.south);
    \draw[->] ([xshift=3em] bert.north) -- ([xshift=3em]linear.south);
    \draw[->] ([xshift=6em] bert.north) -- ([xshift=6em]linear.south);
    \draw[->] ([xshift=-6em] bert.north) -- ([xshift=-6em]linear.south);
    \draw[->] ([xshift=7em] bert.north) -- ([xshift=7em]linear.south);
    \draw[->] ([xshift=-7em] bert.north) -- ([xshift=-7em]linear.south);
    
    \draw[->] ([xshift=5em] linear.north) -- ([xshift=5em]sig.south);
    \draw[->] ([xshift=4em] linear.north) -- ([xshift=4em]sig.south);
    \draw[->] ([xshift=-5em] linear.north) -- ([xshift=-5em]sig.south);
    \draw[->] ([xshift=-4em] linear.north) -- ([xshift=-4em]sig.south);
    \draw[->] ([xshift=-3em] linear.north) -- ([xshift=-3em]sig.south);
    \draw[->] ([xshift=3em] linear.north) -- ([xshift=3em]sig.south);
    \draw[->] ([xshift=6em] linear.north) -- ([xshift=6em]sig.south);
    \draw[->] ([xshift=-6em] linear.north) -- ([xshift=-6em]sig.south);
    \draw[->] ([xshift=7em] linear.north) -- ([xshift=7em]sig.south);
    \draw[->] ([xshift=-7em] linear.north) -- ([xshift=-7em]sig.south);

    \draw[->] ([xshift=5em] sig.north) -- ([xshift=5em]output.south);
    \draw[->] ([xshift=4em] sig.north) -- ([xshift=4em]output.south);
    \draw[->] ([xshift=-5em] sig.north) -- ([xshift=-5em]output.south);
    \draw[->] ([xshift=-4em] sig.north) -- ([xshift=-4em]output.south);
    \draw[->] ([xshift=-3em] sig.north) -- ([xshift=-3em]output.south);
    \draw[->] ([xshift=3em] sig.north) -- ([xshift=3em]output.south);
    \draw[->] ([xshift=6em] sig.north) -- ([xshift=6em]output.south);
    \draw[->] ([xshift=-6em] sig.north) -- ([xshift=-6em]output.south);
    \draw[->] ([xshift=7em] sig.north) -- ([xshift=7em]output.south);
    \draw[->] ([xshift=-7em] sig.north) -- ([xshift=-7em]output.south);

\end{tikzpicture}

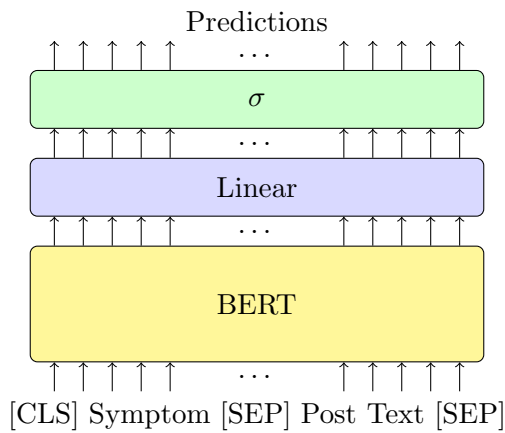
\captionof{figure}{The attribute extraction model architecture.}
\label{model_attribute}
\end{center}

\subsubsection{Start-End Post-Processing}
\label{postprocessing}
To generate the predictions for the Start-End method, we consider each combination of tokens $t_i$, $t_j$ with $i \leq j$ and compute $$p_{i,j}^\text{range} = \frac{p^s_i + p^e_j}{2}.$$
We construct the predicted attributes as all token sequences $t_i, \dots, t_j$ where $p_{i,j}^\text{range} > \tau_\text{startend} = 0.7$. To avoid predicting intersecting attributes, we also require

\begin{gather*}
    \frac{2}{3} \cdot p_{i,j}^\text{range} > p^s_k \text{ and } \\ \frac{2}{3} \cdot p_{i,j}^\text{range} > p^e_k \quad \text{for all } k \in \{i+1, \dots, j-1\}.
\end{gather*}
Additionally, we limit the maximum number of tokens to avoid long predictions due to mismatched start and end tokens.

\subsection{Training}
Following HuggingFace's question-answering examples,\footnote{\url{https://github.com/huggingface/transformers/tree/master/examples/question-answering}} we use the Adam optimizer (\citealt{DBLP:journals/corr/KingmaB14}) with learning rate $3\text{e-}5$ and batch size $32$.\footnote{We accumulate batches to conserve GPU memory.} We train for $40$ epochs, and run evaluation every $50$ steps, with early stopping using the validation set. We use the default BERT dropout configurations for all symptom classification models and increase them to $0.2$ for the attribute extraction models. 

\clearpage

\section{Symptom Hierarchy}\label{symptomtree}
\begin{center}
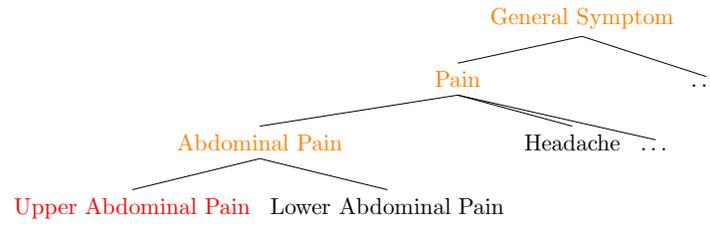

    \centering
   \begin{tikzpicture}[scale=0.8]
  \Tree[.\textcolor{orange}{General~Symptom}
        [.\textcolor{orange}{Pain}
            [.\textcolor{orange}{Abdominal~Pain}
                \textcolor{red}{Upper~Abdominal~Pain}
                Lower~Abdominal~Pain
            ]
            Headache
            $\dots$
        ]
        $\dots$
    ]
    \end{tikzpicture}
    \captionof{figure}{A patient that has the symptom ``Upper Abdominal Pain'' semantically also has all of its parent symptoms; ``Abdominal Pain'', ``Pain'' and ``General Symptom''.}
    \label{parents}
\end{center}

\clearpage
\section{Generalization Ability}\label{app:generalization}
In Table \ref{augvsdefault} we observe that the symptom query model seems to perform better on symptoms that appear less frequently in the train set. The difference decreases as symptom frequency increases. We hypothesize that the difference can be attributed to better generalization ability of the $\text{BERT}_\text{SQ}$ architecture.
\begin{center}
    {\small
        \begin{tabular}{lrrrr}
            \toprule
                                                          \specialcellleft{Symptom Name\\(translated)} & \specialcell{Symptom Frequency \\ (in train set)} &   \specialcell{$\text{BERT}_\text{SQ}$ CL+AD} &   \specialcell{$\text{BERT}_\text{SQ}$ CL} &  \specialcell{Difference} \\
            \midrule
                                                      Bloating & 7 &  $0.00$&      $1.00$& $-1$ \\
                                  \multicolumn{5}{c}{\vdots } \\
                                                Pain &  48 & $0.92$&     $0.86$& $0.06$ \\
                                                          Abdominal Pain &  43 & $1.00$&      $0.91$& $0.09$ \\
                                Nausea & 33 & $0.80$&      $0.57$& $0.23$\\
                                 Hematochezia & 15 & $1.00$&      $0.75$& $0.25$\\
                                          Upper abdominal pain & 23  & $1.00$&      $0.67$& $0.33$\\
                                                      Vomiting & 13 & $0.67$&      $0.00$& $0.67$\\
                                                     Respiratory Complaints & 0 & $1.00$&      $0.00$& $1.00$\\
                                                     Neurological Complaints & 0 & $1.00$&      $0.00$& $1.00$\\
                                                     Agitation & 1 & $1.00$&      $0.00$& $1.00$\\
            \bottomrule
        \end{tabular}
    }
    \centering
    \captionof{table}{Per symptom F1 scores on the held out test set for the BERT Symptom Query model with curriculum learning, and the model trained on the augmented descriptions dataset with curriculum learning. Only symptoms with different F1 scores are presented.}
    \label{augvsdefault}
\end{center}

\clearpage
\section{Attribute Extraction Metrics}\label{app:more_results}
\begin{center}
\begin{tabular}{lrrrrr}
        \toprule
        Method            & Location            & Description       & Time           & Frequency      & Action        \\
        \midrule
        $\text{BERT}_\text{start\_end}$         & $0.60$     & $0.34$    & $0.63$    & $0.26$    & $0.11$\\
        $\text{BERT}_\text{start\_end}$ general & $0.68$    & $0.34$    & $0.27$    & $0.17$    & $0.34$\\
        $\text{BERT}_\text{contiguous}$         & $0.59$    & $0.44$    & $0.44$    & $0.49$    & $0.33$\\
        $\text{BERT}_\text{contiguous}$ general & $0.53$    & $0.29$    & $0.50$     & $0.47$    & $0.76$\\
        \bottomrule
    \end{tabular}
    \captionof{table}{Attribute extraction results on the held out test set: token-wise recall based on the constructed predictions}
    \begin{tabular}{lrrrrr}
        \toprule
        Method            & Location            & Description           & Time      & Frequency & Action        \\
        \midrule
        $\text{BERT}_\text{start\_end}$         & $0.52$    & $0.20$    & $0.31$    & $0.15$    & $0.13$\\
        $\text{BERT}_\text{start\_end}$ general & $0.72$    & $0.54$    & $0.31$    & $0.23$    & $0.22$\\
        $\text{BERT}_\text{contiguous}$         & $0.69$    & $0.42$    & $0.24$    & $0.28$    & $0.16$\\
        $\text{BERT}_\text{contiguous}$ general & $0.86$    & $0.44$    & $0.26$    & $0.38$    & $0.20$\\
        \bottomrule
    \end{tabular}
    \captionof{table}{Attribute extraction results on the held out test set: token-wise precision based on the constructed predictions}

\end{center}
\clearpage

\section{Example Predictions}\label{app:example_preds}

\subsection{Start End general model}
\label{start_end_single_prediction}
    Symptom: Abdominelle Schmerzen (abdominal pain)
\\
\\
Hallo, wie der Titel schon sagt, ich habe \textbf{ständig Bauchschmerzen, meistens} \textbf{auf der linken Seite}, \textbf{neben dem Bauchnabel}, da tut es auch immer weh, \textbf{wenn man draufdrückt}. War damit mehrfach beim Arzt und mir wurden Protonenhemmer verschrieben, die aber überhaupt nicht geholfen haben. Mein Arzt meint, ich habe vielleicht H. pylori und nun habe ich die Wahl zwischen einem H. pylori Atemtest, den ich selber zahlen muss oder einer Endoskopie, wobei der Arzt meint, dass das sehr unangenehm ist. Ich weiss nicht so recht, mir kommt das alles etwas seltsam vor, aber konnte mit meinem Arzt nicht so richtig darüber reden, weil ich eh schon Angst habe, dass er meint, ich spinne nur rum. H. pylori befällt doch den Magen und den Zwölffingerdarm, ist das nicht alles viel höher als neben dem Bauchnabel? Wird nicht erstmal ein Bluttest gemacht oder so etwas, bevor man gleich die schweren Geschütze auffährt? [...]
%Kann mir vielleicht irgendwer weiterhelfen? Habe mich nun spontan für den Atemtest entschieden, weil ich vor einer Endoskopie echt bammel habe, aber so oder so folgt die ja dann wenn kein H. pylori festgestellt wird, von daher ist es auch quatsch oder? Ist so eine Endoskopie wirklich so schlimm, man kann doch da auch was zur Beruhigung bekommen, ist ja schon ne ganze Stange Geld dieser pylori Test (um die 80 Euro).

\begin{center}
\begin{tabular}{lrr}
	\toprule
Attribute & Probabililty & Predicted Text       \\
\midrule
Location & 0.9989      & auf der linken Seite \\
 & 0.9972      & neben dem Bauchnabel                \\
\hline
Frequency & 0.9990      & ständig Bauchschmerzen, meistens \\
\hline
Action & 0.9975      & wenn man draufdrückt \\
\bottomrule
\end{tabular}
\end{center}

\subsection{Contiguous general model}
\label{contiguous_single_prediction}
    Symptom: Abdominelle Schmerzen (abdominal pain)
    \\
    \\
Hallo, wie der Titel schon sagt, ich habe \textbf{ständig Bauch}schmerzen, \textbf{meistens auf der linken Seite, neben dem Bauchnabel}, da tut es auch \textbf{immer} weh, \textbf{wenn man draufdrückt}. War damit mehrfach beim Arzt und mir wurden Protonenhemmer verschrieben, die aber überhaupt nicht geholfen haben. Mein Arzt meint, ich habe vielleicht H. pylori und nun habe ich die Wahl zwischen einem H. pylori Atemtest, den ich selber zahlen muss oder einer Endoskopie, wobei der Arzt meint, dass das sehr unangenehm ist. Ich weiss nicht so recht, mir kommt das alles etwas seltsam vor, aber konnte mit meinem Arzt nicht so richtig darüber reden, weil ich eh schon Angst habe, dass er meint, ich spinne nur rum. H. pylori befällt doch den Magen und den Zwölffingerdarm, ist das nicht alles viel höher als \textbf{neben dem Bauchnabel}? Wird nicht erstmal ein Bluttest gemacht oder so etwas, bevor man gleich die schweren Geschütze auffährt? [...]
%Kann mir vielleicht irgendwer weiterhelfen ? Habe mich nun spontan für den Atemtest entschieden , weil ich vor einer Endoskopie echt bammel habe , aber so oder so folgt die ja dann wenn kein H . pylori festgestellt wird , von daher ist es auch quatsch oder ? Ist so eine Endoskopie wirklich so schlimm , man kann doch da auch was zur Beruhigung bekommen , ist ja schon ne ganze Stange Geld dieser pylori Test ( um die 80 Euro ) . . .

\begin{center}
\begin{tabular}{lrr}
	\toprule
Entity & Probabililty & Predicted Text       \\
\midrule
Location & 0.9990      & Bauch                \\
 & 0.9994      & auf der linken Seite, neben dem Bauchabel \\
  & 0.9990      & neben dem Bauchnabel                \\
\hline
Frequency & 0.9997      & ständig \\
  & 0.9960      & meistens \\
  & 0.9996      & immer \\
\hline
Action & 0.9997      & wenn man draufdrückt \\
\bottomrule
\end{tabular}
\end{center}

\end{document}